\documentclass[preprint,5p,times,twocolumn]{elsarticle}
\pdfoutput=1
\usepackage{amsmath,amsfonts}
\usepackage{amssymb}
\usepackage{mathrsfs}
\usepackage{lipsum}
\usepackage{graphicx}
\usepackage{textcomp}
\usepackage{multirow}
\usepackage{multicol}
\usepackage{algpseudocode}
\usepackage[switch]{lineno} 

\begin{document}

\begin{frontmatter}

\title{Feature-oriented Deep Learning Framework for Pulmonary Cone-beam CT (CBCT) Enhancement with Multi-task Customized Perceptual Loss}

\author[a,1]{Jiarui Zhu }
\author[b]{Weixing Chen }
\author[c]{Hongfei Sun}
\author[a]{Shaohua Zhi}
\author[a]{Mayang Zhao}
\author[d]{Lam Yu Lap}
\author[e]{Zihan Zhou}
\author[f]{Tao Peng}

\author[]{Jing Cai$ ^{a,*}$}
\author[]{Ge Ren$ ^{a,g,*}$}

\affiliation[a]{organization={Department of Health Technology and Informatics, The Hong Kong Polytechnic University},
            postcode={999077}, 
            state={Hong Kong},
            country={China}}
\affiliation[b]{organization={School of Computer Science and Engineering, Sun Yat-Sen University},
            postcode={510006}, 
            state={Guangzhou},
            country={China}}
\affiliation[c]{organization={Department of Radiation Oncology, Xijing Hospital, Fourth Military Medical University},
            postcode={710032}, 
            state={Xian},
            country={China}}
\affiliation[d]{organization={Department of Clinical Oncology, Queen Mary Hospital},
            postcode={999077}, 
            state={Hong Kong},
            country={China}}
\affiliation[e]{organization={Department of Radiation Oncology, Jinshazhou Hospital of Guangzhou University of Chinese Medicine},
            postcode={510168}, 
            state={Guangzhou\affiliation[f]{organization={School of Computer Science and Engineering, Sun Yat-Sen University},
            postcode={510006}, 
            state={Guangzhou},
            country={China}}},
            country={China}}
\affiliation[f]{organization={School of Future Science and Engineering, Soochow University},
            postcode={215299}, 
            state={Suzhou},
            country={China}}
\affiliation[g]{organization={Research Institute for Intelligent Wearable Systems, The Hong Kong Polytechnic University},
            postcode={999077}, 
            state={Hong Kong},
            country={China}}




     
\begin{abstract}
Cone-beam computed tomography (CBCT) is routinely collected during image-guided radiation therapy (IGRT) to provide updated patient anatomy information for cancer treatments. However, CBCT images often suffer from streaking artifacts and noise caused by under-rate sampling projections and low-dose exposure, resulting in low clarity and information loss. 
While recent deep learning-based CBCT enhancement methods have shown promising results in suppressing artifacts, they have limited performance on preserving anatomical details since conventional pixel-to-pixel loss functions are incapable of describing detailed anatomy.
To address this issue, we propose a novel feature-oriented deep learning framework that translates low-quality CBCT images into high-quality CT-like imaging via a multi-task customized feature-to-feature perceptual loss function. 
The framework comprises two main components: a multi-task learning feature-selection network(MTFS-Net) for customizing the perceptual loss function; and a CBCT-to-CT translation network guided by feature-to-feature perceptual loss, which uses advanced generative models such as U-Net, GAN and CycleGAN. 
Our experiments showed that the proposed framework can generate synthesized CT (sCT) images for the lung that achieved a high similarity to CT images, with an average SSIM index of 0.9869 and an average PSNR index of 39.9621. 
The sCT images also achieved visually pleasing performance with effective artifacts suppression, noise reduction, and distinctive anatomical details preservation. 
Functional imaging tests further demonstrated the pulmonary texture correction performance of the sCT images.
Comparison experiments with pixel-to-pixel loss also showed that the proposed perceptual loss significantly enhances the performance of involved generative models. 
Our experiment results indicate that the proposed framework outperforms the state-of-the-art models for pulmonary CBCT enhancement. This framework holds great promise for generating high-quality anatomical imaging from CBCT that is suitable for various clinical applications, such as CBCT-based adaptive radiation therapy, functional imaging synthesis, and radiomics analysis.
\end{abstract}


\end{frontmatter}


\section{Introduction}
\label{introduction}

\footnotetext[1]{Github link:https://github.com/zhujiarui42/CFP-Loss}
\footnotetext[2]{Contact Email:zhujiarui42@gmail.com}

\par Cone-beam computed tomography (CBCT) images are reconstructed from sparse-view cone-beam projections\cite{cbctdefinition}. This volumetric cone-beam acquisition technique enables the CBCT imaging system to obtain patient volume information quickly with low radiation exposure, making it widely used to provide up-to-date on-board images during image-guided radiation therapy (IGRT)\cite{onboardcbct}. This on-board CBCT imaging is clinically utilized for patient setup and treatment monitoring, particularly for cancer-based patient positioning\cite{positioning}. However, CBCT images still suffer from severe streaking artifacts, noise, poor preservation of anatomical details, and unreliable Hounsfield unit (HU) values\cite{artifactsreview}, thereby limiting its potential clinical applications, such as CT-based adaptive radiation therapy (ART) \cite{art}, functional imaging\cite{perfusion}, and radiomics analysis\cite{radiomics}. Therefore, there is a pressing need for a novel technique to improve the quality of CBCT imaging, which would be highly valuable and meaningful.

\par To overcome these drawbacks, considerable efforts have been dedicated to enhancing the image quality of CBCT. Traditional methods for reducing scatter photons in two-dimensional (2D) projection images involve the use of anti-scatter grid hardware suppression and scatter deconvolution estimation\cite{convenre1,convenre2,convenre3}. However, these methods only offer moderate artifact suppression and incur heavy computational costs. Alternatively, iterative projection-reconstruction methods have been developed to enhance CBCT image quality by artifact compensation and information interpolation. These methods employ a range of spatiotemporal regularizers coupled with mathematical constraints \cite{mathcons2}, or leverage prior knowledge \cite{mathconsadd,cycn,recondp}. Iterative projection-reconstruction methods are effective in compensating for motion artifacts and suppressing streaking artifacts. Nevertheless, these methods may compromise the preservation of anatomical details and typically require access to raw projection data.

\par Recently, deep learning (DL)-based image translation models have gained significant popularity. As the CBCT-paired CT is commonly collected for the treatment planning stage in radiation therapy (RT), many researchers have focused on developing novel CBCT-to-CT translation DL models to enhance CBCT image quality to CT image level. U-Net was firstly applied to minimize the pixel-to-pixel difference between the enhanced CBCT and the corresponding CT\cite{mathcons1,uneten}. Although U-Net models have fast and stable convergence, the detail structure recovery is still limited. To improve the model's tunability, GAN models with pixel-to-pixel adversarial loss were used to improve the model performance\cite{ganen}. Another challenge for this translation is the patient position mismatch gap between CBCT scan and CT scan, which degraded the label accuracy during the learning process. To address this problem, Cycle GAN models with cycle-consistency loss were introduced and generated sCT images with satisfying artifacts suppression within pelvic\cite{cycleganpelvic}, neck\cite{cycleganneck}, brain\cite{cycleganbrain} and prostate\cite{cycleganprostate} regions. DL-based CBCT enhancement models have a promising future, as they offer superior advantages of real-time processing speed and affluent data sources for consistent updating.

\par Although DL-based direct image translation methods have brought significant breakthroughs, there remain several unsolved challenges. Particularly in organ regions with abundant textual details, such as lung regions, the local anatomical details are prone to loss after enhancement due to the mismatched nature of the corresponding CT image\cite{cycleganlung,cycleganlung2}. The anatomical misalignments between CBCT and CT images are entangled with spatial displacement caused by the acquisition time interval between CBCT and CT scans, which degrades the reliance of the target CT images. Modern registration techniques are far from perfect in solving spatial displacement\cite{regisen1,regisen2} because the entangled spatial displacement and anatomical misalignment are hard to distinguish. Although unpaired CycleGAN models partially alleviate the misalignment by forcing a bidirectional mapping and achieved better performance, anatomical details remain poorly preserved.

\par One essential reason for this issue lies in the defective optimization object, as a normal pixel-to-pixel loss function is limited to pixel-based location and fails to describe detailed anatomical information from CT target. As a result, all partial mismatched anatomy is identified as unwanted information and eliminated from CBCT inputs during the iterative optimization process. To generate enhanced CBCT images with well-preserved anatomy, an optimization object with sophisticated high-dimensional descriptive ability is highly demanded. It should be able to capture the global and subtle anatomical features, which can be used for detailed anatomical texture analysis.

\par In this study, we proposed to develop a novel deep learning framework that replaces the conventional pixel-to-pixel loss with a high-dimensional feature-to-feature perceptual loss. Our framework consists of two main components: a multi-task feature-selection network and a CBCT-to-CT translation network. The feature-selection network pre-trains a customized autoencoder that serves as a building block for the perceptual loss function in the translation network. The autoencoder is trained with three unique tasks that share the same backbone structure, and the loss functions from these tasks are combined through a gradnorm regularization technique to refine the autoencoder further. The CBCT-to-CT translation network utilizes the customized autoencoder to generate high-quality synthesized CT images from CBCT scans. The translation network employs the perceptual loss between the synthesized CT images and paired CT images to guide the optimization process and to ensure that the fine anatomical details are well-preserved. Our proposed deep learning framework represents a significant improvement over existing models, as it enables us to generate sCT images with well-preserved anatomy and high image quality.

The main contributions of this study can be summarized as follows:
\begin{enumerate}
\item 
We propose a novel feature-oriented deep learning framework that is specifically designed for enhancing the quality of CBCT images. Our approach leverages a multi-task feature-selection network for feature-to-feature perceptual loss build-up in addition to utilizing novel generative models for image translation. Our framework effectively preserves fine anatomical details from CBCT images and transfers the noise-clear trait from CT targets to high-quality synthesized CT (sCT) imaging results.
\item
We propose a unique multi-task neural network with three subtasks and gradnorm regularization to customize a perceptual loss function that can extract features with mixed desired traits from CBCT imaging. The loss function integrates traits from three subtasks, including CT self-recovery task, CBCT-to-CT dual-pyramid registration task, and CBCT-or-CT binary classification task. Compared to the pixel-to-pixel loss function, the proposed feature-to-feature perceptual loss function demonstrates a stronger feature extraction ability for more representative features and further improves sCT imaging quality.
\item 
We provide extensive experimental evidence to demonstrate the effectiveness of our proposed multi-task customized perceptual loss function. The results clearly indicate that our perceptual loss function significantly enhances the performance of all existing generative models for CBCT-to-CT translation, including U-Net, GAN, and the state-of-the-art CycleGAN models.

\end{enumerate}

The remaining sections of this paper are organized as follows:
In Section II, we provide a comprehensive review of related studies on image-to-image translation, perceptual loss, and multi-task learning. Section III presents a detailed explanation of the architecture employed in our proposed framework. In Section IV, we discuss the experimental materials, evaluate our model, and present the obtained experimental results. In Section V , we discuss implications, limitations, future work, and key contributions of our research. Finally, in Section VI, we summarize our work and draw a conclusion.

\section{Related Work}

\subsection {Image-to-image Translation}
Image-to-image translation is a widely used technique in the field of medical imaging, which involves mapping an image from a source modality to a target modality\cite{image2imagereview}. This technique finds applications in various areas, including noise reduction, super-resolution, image synthesis, and reconstruction. However, a significant challenge in image-to-image translation is the ill-posed nature, resulting from the non-unique reverse mapping from the target to the source.
To address this limitation, researchers have developed comprehensive loss functions that consider the relevant internal relationships and capture subtle image details.
For example, Richardson et al. proposed a Context-Aware GAN with an image-gradient-difference-based loss function for CT-to-MR translation\cite{contextgan}. Elad et al. introduced a StyleGAN with a pixel-to-latent loss function that achieved robust generic translation results\cite{stylegan}. Devavrat et al. employed a segmentation-to-real loss function to regularize contrastive unsupervised learning for unpaired simulated-ultrasound to real-ultrasound translation\cite{contrastiveus}. These studies highlight the significance of designing an appropriate loss function to enhance the predictive performance of the network. Inspired by these works, we propose a customized feature-oriented perceptual loss function through a multi-task network that specifically considers the traits of the CBCT-to-CT translation task.

\subsection {Perceptual Loss}
Perceptual loss is a loss function commonly utilized in image-to-image translation problems, which quantifies the dissimilarity between the high-dimensional feature maps extracted from the network output and the target. This is achieved by employing a pre-trained perceptual loss network and calculating the difference typically using Euclidean distance.

The original perceptual loss network, based on VGG-16, was trained on the ImageNet dataset and initially applied to mismatched natural image super-resolution and style-transfer tasks\cite{li2016}. Subsequently, researchers have incorporated the original perceptual loss function into various applications with impressive results. For example, Yang et al. employed the original perceptual loss function in WGAN for low-dose CT denoising, achieving remarkable performance\cite{tmi2018}. Georgios et al. combined CycleGAN with the original perceptual loss function for spectral domain optical coherence tomography classification\cite{cycleperceptual}. Ran et al. utilized the original perceptual loss along with MSE loss to train a residual encoder-decoder WGAN for MRI denoising \cite{mridenoisingperceptual}. Li et al. initialized the parameters of a VGG-14 loss network by training a self-supervised NDCT recovery network, and the resulting perceptual loss achieved state-of-the-art performance for LDCT denoising with both attention U-Net and WGAN\cite{tmi2020}.

Moreover, perceptual loss is not limited to image-to-image translation tasks and has also proven effective in image classification and segmentation. Cheng et al. constructed a perceptual loss function based on non-local spectral and structural similarities, which yielded excellent performance for multi-spectral and panchromatic image classification\cite{adapativeperceptual}. High-dimensional feature maps capture essential features and encode a manifold that encompasses key information. A well-designed perceptual loss has the potential to comprehensively describe the correlation between the network output and the target, accounting for anatomical differences and spatial misalignment.

\subsection {Multi-task Learning}

Multi-task learning (MTL) is a powerful approach in machine learning that involves training multiple related tasks simultaneously by sharing feature representations and optimizing a combined loss function\cite{mtlreviewml2018}. In the field of medical imaging, MTL using deep learning has shown significant potential in improving model accuracy and robustness by leveraging feature selection and traits transfer from subtasks\cite{mtlreviewdp2017}.

Various studies in medical imaging have demonstrated the effectiveness of MTL. For instance, Antari et al. combined tasks such as mammography segmentation, classification, and object detection\cite{mtla2}. Kyung et al. integrated head CT classification, segmentation, and reconstruction as three subtasks\cite{mtla3}. These works have shown that models trained using MTL outperform individually trained subtasks by benefiting from better feature selection.

Furthermore, other studies have showcased the advantages of MTL for specific medical applications. Boutilon et al. combined multi-scale contrastive classification and multi-joint anatomical segmentation for the clinical diagnosis of the pediatric musculoskeletal system\cite{mtla4}. Lei et al. combined brain functional imaging synthesis and brain structural imaging synthesis for early diagnosis and intervention of mild cognitive impairment\cite{mtla5}. These works demonstrate that MTL enables the transfer of traits among various subtasks, resulting in the development of a more comprehensive encoder.

In general, the MTL paradigm maximizes the utilization of information from subtasks, mitigating overfitting and yielding a more robust model with superior accuracy and generalization performance. By sharing the same encoder and optimizing a combined loss function, MTL imposes strong constraints on feature selection, leading to the refinement of feature maps and the facilitation of desired traits transfer. Building upon these insights, we propose a multi-task learning structure that incorporates tasks containing desired traits to construct a relevant perceptual loss function. By leveraging MTL and designing a customized loss function that considers the traits of the tasks, we can sensitively capture subtle details and achieve improved performance in image-to-image translation tasks.

\section{Methodology}
\begin{figure*}
  \includegraphics[width=7in]{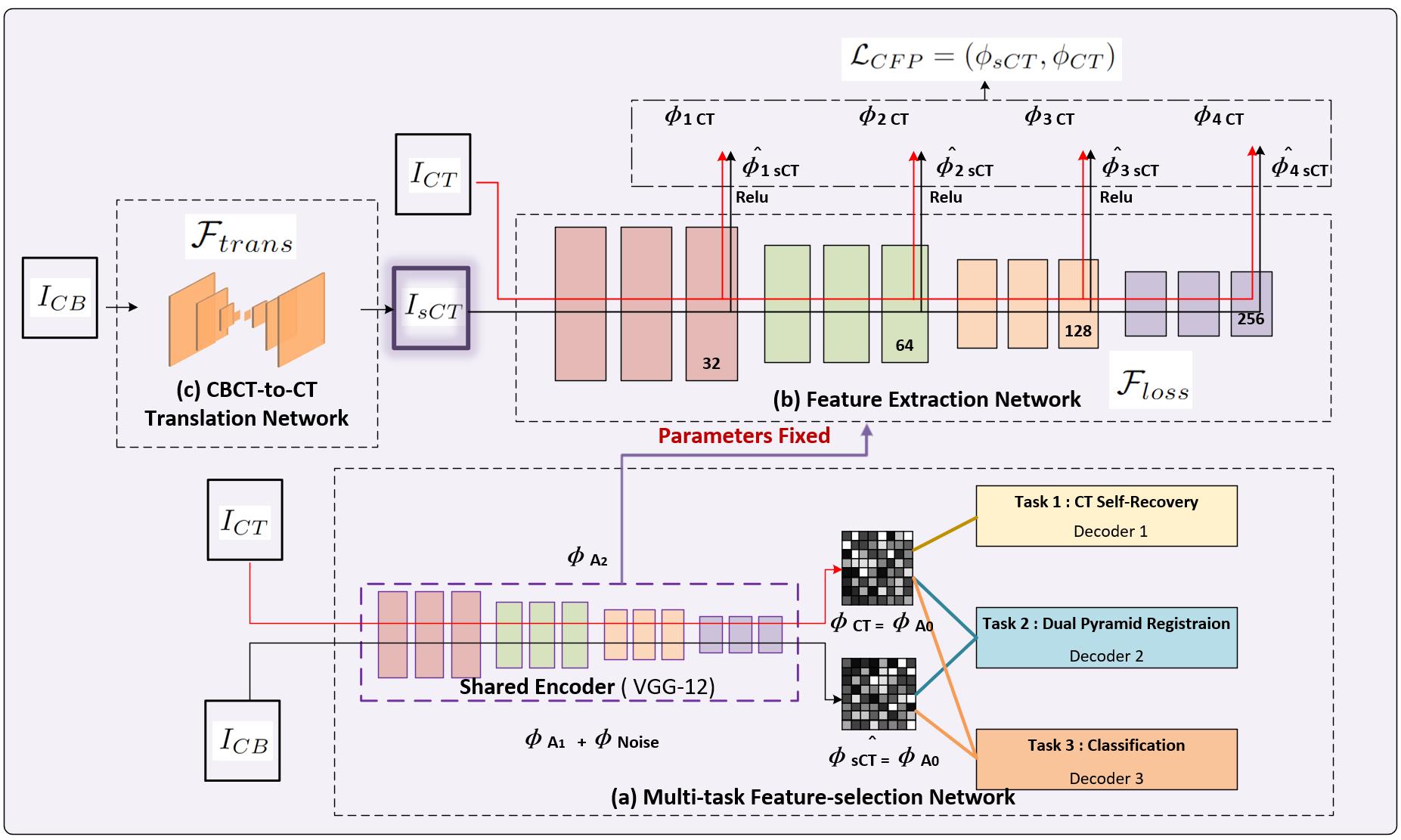}
  \caption{The overall architecture of our proposed framework.(a)indicates our multi-task feature-selection network (b)indicates our feature extraction network (c)indicates our CBCT-to-PlanCT translation network.}\label{overall}
\end{figure*}

\par The CBCT-to-CT translation problem can be formulated as a cross-modality translation model in the image spatial domain. Let $I_{CB}\in\mathbb{R}^{H*W}$ denote a CBCT imaging, where H, W represent height and width of the image.  Let $I_{CT}\in\mathbb{R}^{H*W}$denote a CT imaging. The goal of the translation is to seek a function $\mathcal{F}_{trans}$ that maps a CBCT image to a CT image, and a synthesized CT(sCT) ${I}_{sCT}$ is generated from CBCT ${I}_{CB}$ by $\mathcal{F}_{trans}$ : 
\begin{equation}
    \mathcal{F}_{trans}(I_{CB};\Theta) = {I}_{sCT}
\end{equation}

\par where $\Theta$ refers to weighing parameters of $\mathcal{F}_{trans}$.

\par And by adopting a CNN model to generate ${I}_{sCT}$ from  ${I}_{CB}$ , the translation is formulated as an optimization problem, and the optimization object is commonly built-up as a loss function $\mathcal{L}$:

\begin{equation}
    \hat{\Theta}= arg\mathop{min}\limits_{\mathcal{L}}\mathcal{L}[\mathcal{F}_{trans}(I_{CB};\Theta),I_{CT}]
\end{equation}

\par Our approach to solve the translation problem adopts a feature-oriented framework, which comprises two main components. The first component is CBCT-to-CT translation division, which extracts key features from CBCT($I_{CB}$) and generates an sCT($I_{sCT}$) that closely resembles the CT($I_{CT}$) using novel generative CNN models designed specifically for the modality difference. The second component is the feature-selection division, which customizes the modality difference between CBCT and CT through multl-task feature-selection.

\par To achieve this, we adopt a pre-trained autoencoder, denoted as $\mathcal{F}_{loss}$, to extract compressed feature maps $\phi_{CB}$ and $\phi_{CT}$ from CBCT($I_{CB}$) and CT($I_{CT}$), respectively. We assume that the feature map $\phi_{CB}$ can be decomposed as the sum of anatomical information $\phi_{A_1}$ from CBCT and irrelevant information $\phi_{noise}$, which includes artifacts and noise caused by cone-beam reconstruction and low-dose exposure. 

\par Similarly, the feature map $\phi_{CT}$ represents the anatomical information from CT. Despite the anatomical differences between CBCT and CT, our goal is to identify the corresponding features $\phi_{A_0}$ that encode the same anatomical information in both domains(i.e., $\phi_{A_0} = \phi_{A_1} \cap \phi_{A_2})$. In addition to identifying $\phi_{A_0}$, we aim to transfer the noise-free trait from $\phi_{CT}$ to $ \phi_{CB}$ to remove the irrelevant information (i.e.,$\phi_{noise}$) from the CBCT feature map. Ultimately, we propose to customize a feature-refined autoencoder (FAE) as the loss net $\mathcal{F}_{loss}$, which can extract the desired feature map $\phi_{A_0}$ from $I_{CT}$:

\begin{equation}
    \phi_{A_0(I_{CT})} = \mathcal{F}_{loss}(I_{CT})
\end{equation}

\par and the same loss net is used to extract $\phi_{A_0}$ from ${I}_{sCT} = \mathcal{F}_{trans}(I_{CB})$ :

\begin{equation}
    \phi_{A_0(I_{sCT})} = \mathcal{F}_{loss}(I_{sCT})
\end{equation}

And then a feature-oriented perceptual loss function $\mathcal{L}_{perceptual}(I_{CT}, I_{sCT})$ was formulated to calculate a detailed high-dimensional modality difference:

\begin{equation}
    \mathcal{L}_{perceptual}(I_{CT},I_{sCT}) = 
    \mathcal{L}(\phi_{(I_{CT})},\phi_{(I_{sCT})})
\end{equation}

\par Fig.~\ref{overall} illustrates the architecture of our proposed feature-oriented framework. Our scheme comprises a 3-task multi-task feature-selection network(MTFS-Net) to customize the FAE. As is shown in Fig.~\ref{overall}(a), the 3 tasks include a CT-to-CT self-recovery task 1 for extracting $\phi_{A_2}$, a dual-pyramid registration task 2 for locating feature correlations between CBCT and CT to locate an intersection between $\phi_{A_1} + \phi_{noise}$ and $\phi_{A_2}$, and a binary classification task 3 for disentangling between $\phi_{A_1} + \phi_{noise}$ from $I_{CB}$ and $\phi_{A_2}$ from $I_{CT}$. By jointly training task 1 and task 3, we aim to locate $\phi_{noise}$ from $\phi_{A_1} + \phi_{noise}$ based on $\phi_{A_2}$. By jointly training task 1 and task 2, we aim to refine the same pattern $\phi_{A_0}$ from $\phi_{A_1} + \phi_{noise}$ and $\phi_{A_2}$. And by jointly training all 3 tasks together we aim to enable a feature-selection on the backbone autoencoder to: contain complete $A_2$ information, identify $\phi_{noise}$ information from $\phi_{A_1} + \phi_{noise}$ and locate same $\phi_{A_0}$ style from both $\phi_{A_1}$ and $\phi_{A_2}$.

\par After FAE refinement, as is shown in Fig.~\ref{overall}(c), we adopt a CBCT-to-CT translation network consisting of the parameter-fixed loss net and several optional generative models including U-Net, GAN and CycleGAN. In Fig.~\ref{overall}(b), we use the pre-trained FAE from Fig.~\ref{overall}(a) as the loss net to define the perceptual loss function, and use the perceptual loss to guide the training of the generative models for CBCT-to-CT translation.

 \par The MTFS-Net and CBCT-to-CT translation network are introduced in detail in the following sections.

\subsection{Multi-task Feature-Selection Network}
\begin{figure*}
  \includegraphics[width=7in]{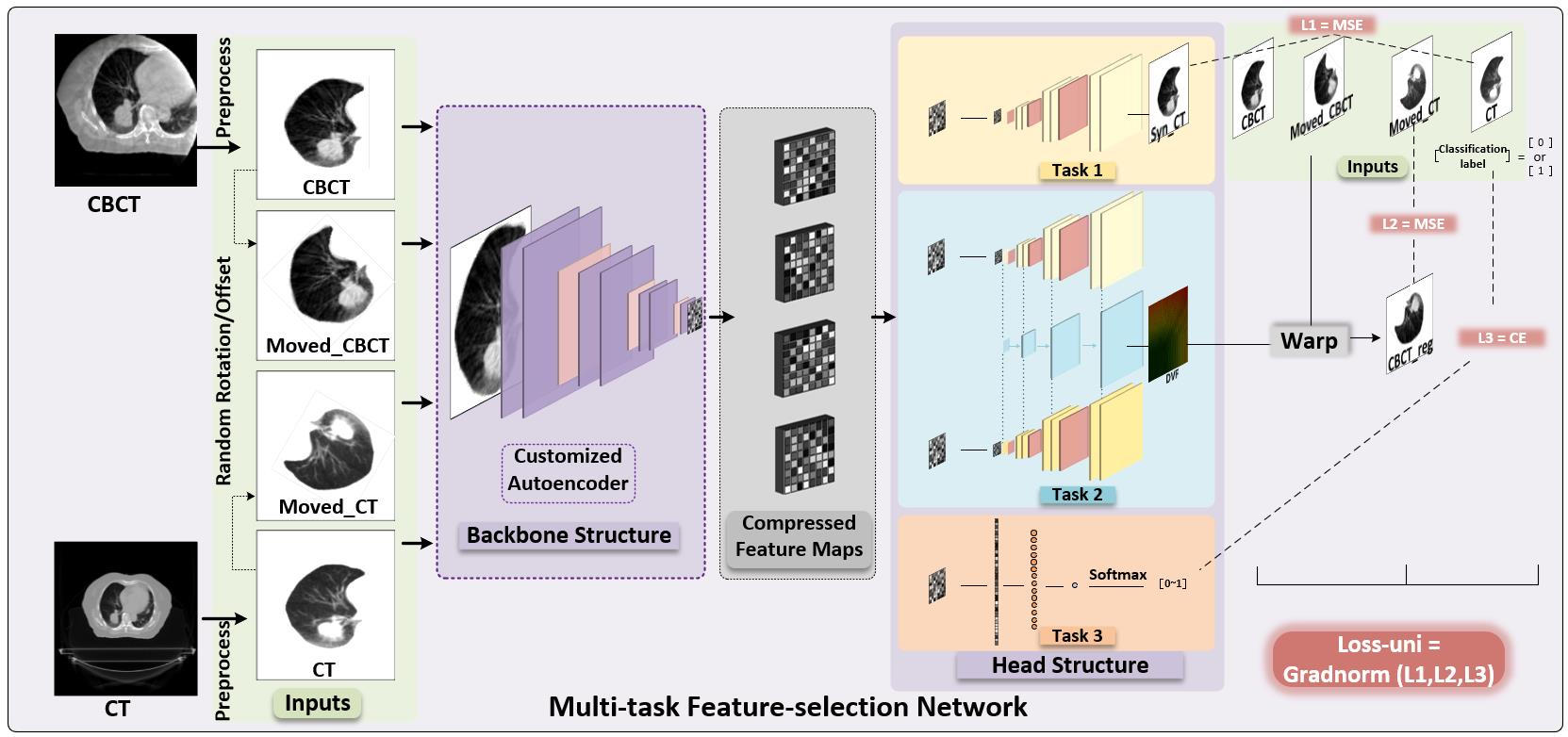}
  \caption{The architecture of the three-task multi-task feature-selection network. The network consists of a commonly shared autoencoder as shared backbone, and three separate subtasks as head structure. The autoencoder is customized by the three subtasks, and the final loss function is untied by three loss functions in a gradnorm form.}\label{mtlnet}
\end{figure*}

Fig.~\ref{mtlnet} shows the architecture of the multi-task feature-selection network, which is based on a “hard-sharing” MTL design. The network has one shared backbone structure and three independent head structures modeling three different tasks. The backbone structure is a deep autoencoder that creates a feature space shared by the three independent head structures. Task 1 aims to extract the manifold from CT, while task 2 selects corresponding features between CBCT and CT domains, and task 3 locates noise and streaking artifacts in the CBCT domain based on the difference between CBCT and CT. These three tasks work together to guide the autoencoder in building a refined feature space with mixed desired traits, which only encodes features containing key information for recovering CT imaging, extracting the same pattern from both CBCT and CT domains, and excluding noise or streaking artifacts.

\subsubsection{Deep Autoencoder Backbone}
Previous research commonly used VGG-like autoencoders as the loss network for perceptual loss function construction. Since that increasing the structure depth enables the autoencoder to further reduce information redundancy and extract more compressed feature maps\cite{autoencoder}, we try to increase the depth of our autoencoder backbone by using skip connections\cite{resnet} so the shared backbone has a 101-layer deep  ResNet architecture. The output size is cautiously set because over-large size may include over-redundant information, yet undersized compressed feature maps may be incapable of containing complete information. As is shown in Fig.~\ref{mtlnet}, after fine adjustment, the autoencoder is designed to include 3 maxpooling layers for downsampling, (31) two-skip residual layers(each contains two sequential 3*3 convolutional layers) and 5 convolutional layers with parameters set to be [(1,32,256,256),(32,64,128,128),(64,128,64,64),(128,256,32,32)] in terms of [(input channels, output channels, H, W)].

\subsubsection{CT Self-recovery Task}
Task 1 is a single-input self-supervised decoder for CT-to-CT recovery with a VGG architecture. Previous works\cite{tmi2018} have proved that a shallow VGG structure is well-qualified for CT self-recovery. As is shown in Fig.~\ref{mtlnet}, we used a VGG-12 structure consisting of 9 convolutional layers with out-channel numbers:256,256,256,128,128,64,64,32,1.

\par For the loss function of self-recovery task1, we utilize mean square error(MSE) function to describe the intensity similarity between the decoder output $\hat{y_1}$, $\hat{y_3}$ and the ground truth ${y}_{CB}$,${y}_{CT}$:

\begin{equation}
\mathscr{Loss}_1 = MSE(\hat{y},y)
\end{equation}

\subsubsection{Dual-pyramid Registration Task}
Task 2 is a dual-input self-supervised decoder generating a deformable vectors field(DVF) for registration between CBCT and CT with random rotation and offset. Registration networks essentially learn reversible pixel-wise or voxel-wise correspondences between two input image domains in the form of DVF\cite{voxelmorph}. Kang et al. proved that DVF decoded from separate feature pyramids dominantly boosts the registration performance\cite{dualpr}. Kang et al. then interpreted that such performance improvement originates from the local anatomical structural details encoded by the feature pyramids. We further infer that the dual-pyramid registration network structure decodes the correlations by selectively encoding strongly correlated features extracted from two input image domains. Therefore, we imply dual-pyramid registration in our multi-task feature-selection network, to select highly-correlated features. These encoded correlations were usually inferred as a structural similarity, but requires further analysis. 

\par We design task 2 in a dual-pyramid registration network architecture. Task 2 generates a deformable vector field(DVF) $DVF = \mathcal{F}_{reg}(I_{moved{\_}CBCT},I_{CT} ; \Theta)$ from dual inputs, where $\Theta$ is the weighing parameters of $\mathcal{F}_{reg}$. $I_{moved{\_}CBCT}$ is warped by the DVF and label and becomes $Y_{CBCT{\_}reg}$. And the loss function of task 2 is the intensity similarity difference between $Y_{CBCT{\_}reg}$ and $I_{CT}$, vice versa:

\begin{equation}
\mathscr{Loss}_2 = MSE(Y_{CBCT{\_}reg},I_{CT}) \quad or \quad MSE ((Y_{CT{\_}reg},I_{CBCT})
\end{equation}

\subsubsection{Classification Task}
Task 3 is a binary classification decoder with a fully-connected neural network structure. Multiple-classification task trained on ImageNet dataset\cite{imagenet} was initially utilized for the original perceptual building-up loss net\cite{li2016}. The classification results are closely related to the feature embedding quality and both shallow structural features and deep semantic features markedly contribute to the classification efficacy. Given enough samples and multiple targets, a classification task will be capable of extracting abundant and distinctive feature maps which may further serve as an ideal perceptual loss net. Meanwhile, given limited targets, a classification task is also able to distinguish key differences between source and targets, in other words, disentangling source content from target style\cite{contrastiveus}. In our case, severe artifacts and noise can be a dominant difference between CBCT source and CT targets. The feature distinguishable trait enables a classification task to identify this dominant noise feature.

\par We apply one fully-connected layer as the decoder for the binary classification task 3 which classifies CBCT and CT by generating the "label-predict". The classification task is trained in an supervised form. The "label-target" set is composed of CBCT and CT images which are labeled as "0" and "1".

The loss function of the classification task 3 is listed as follows:

\begin{equation}
\mathscr{Loss}_3 = CE(label-predict,label-target)
\end{equation}

To normalize the optimizing speed and magnitude of the loss from the three independent tasks, a gradnorm loss regularization strategy\cite{gradnorm} is applied for combining different task losses into one united loss.

\begin{equation}
\mathscr{Loss}_{uni} = Gradnorm(\mathscr{Loss}_1,\mathscr{Loss}_2,\mathscr{Loss}_3)
\end{equation}

\subsection{CBCT-to-CT translation Network}
\begin{figure*}
  \includegraphics[width=7in]{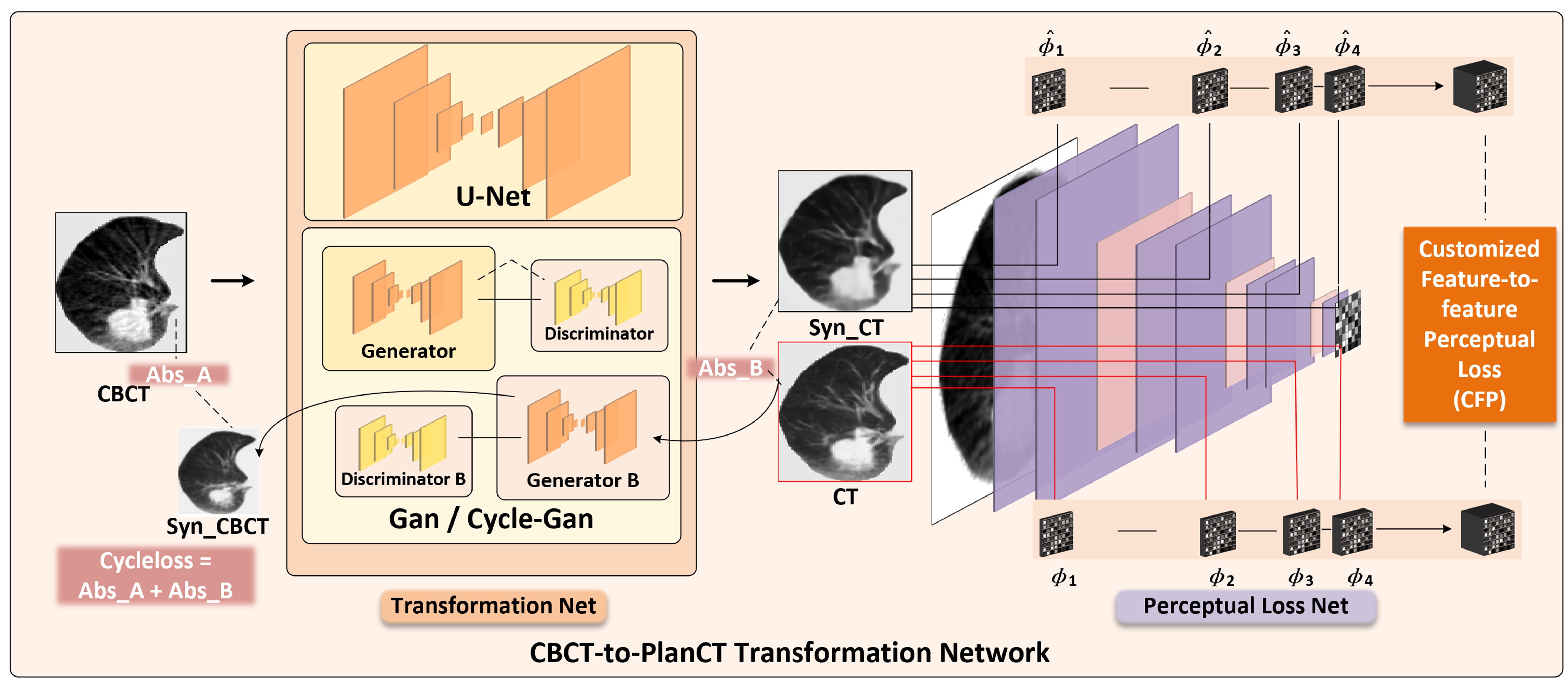}
  \caption{The architecture of the CBCT-to-CT translation network.The network contains a translation net and a perceptual loss net. The translation net employs U-Net, Gan and Cycle-Gan models. The perceptual loss net is the parameter-fixed autoencoder which was pretrained in Fig.~\ref{mtlnet}. And the CFP-loss is calculated by the tensor difference between the feature maps extracted from the synthesized CT and CT by the perceptual loss net.}\label{transnet}
\end{figure*}

Fig.~\ref{transnet} shows the architecture of the CBCT-to-CT translation network, which follows a typical “image translation net + loss net” design \cite{li2016} , consisting of multiple optional generative models as the translation net $F_{trans}(x | \Theta)$ and a pre-trained perceptual loss extraction network $F_{loss}(x | \Theta)$ with fixed parameters, where $x$ is the input and $\Theta$ refers to the network parameters.
Rather than targeting the pixel-to-pixel difference between the pixels of the output image $y = \mathcal{F}_{trans}(x)$ and the ground truth $\hat{y}$, the translation net targets the feature difference between feature representations extracted by the loss net from the output image $\phi_y = \mathcal{F}_{loss}(y)$ and ground truth $\phi_{\hat{y}} = \mathcal{F}_{loss}(\hat{y})$. The feature difference is defined as the perceptual loss $\mathcal{L}_{perceptual}(\phi_y,\phi_{\hat{y}})$.

\subsubsection{Translation Net}
The proposed CBCT-to-CT translation task guided by perceptual loss requires not only preserving low-level information such as textures and contours from CBCT, but also extracting semantic features for distinguishing structural streaking artifacts from textures, identifying noise, and separating $\phi_{A_0}$ from $\phi_{A_1}$. To achieve better translation results, we selected three commonly used generative models, including U-Net, GAN, and CycleGAN, with modifications to better suit our pulmonary CT synthesis case.

The U-Net model is designed in a channel attention residual form(CAR-U-Net). We adopted a U-Net\cite{unet} backbone for CBCT-to-CT translation because the four down-sampling operations allow the extraction of high-level semantic information and their concatenation to four up-sampled layers effectively reserves low-level information. We added residual blocks\cite{resnet} with two-step skip connections(Res-Block) to increase the network depth for higher-level features extraction. And we added classic "squeeze-excite" \cite{senet}channel attention blocks(SE-Block) before each concatenation for adaptive weighing of high-level feature channels.

For the channel attention residual U-Net (CAR-U-Net). Within each sampling level of the "U" shape backbone, we added Res-Blocks with numbers of 2,4,6,8 and we added additional skip connection between the input and the sequential output. Before each concatenation, we added a SE-Block.  

The GAN and CycleGAN models apply same generator and discriminator structure according to previous methods. In our case, we specially replace all the generator loss functions with our proposed perceptual loss function. Especially for CycleGAN, we apply the perceptual loss function $Loss_{perceptual}$ acquired from the FMTL-net, calculate $\phi_{Real}$ from real image $I_{CT}$ or $I_{CB}$ in the backward process of CycleGAN and $\phi_{Syn}$ from fake image generated by generator, and replace the $Loss_{MSE}$ with $Loss_{perceptual}(\phi_{Real},\phi_{Syn})$. We made no modification to the original cycle-consistency loss defined as absolute difference between $\phi_{Real}$ and $\phi_{Syn}$.

\subsubsection{Loss Net}
The pre-trained FAE from the MTFS-Net was directly used as the loss net for the CBCT-to-CT translation network. The loss net parameters remain fixed during the training of the translation net. 

The loss net includes layers of four size-levels including (256*256),(128*128),(64*64) and (32*32) in terms of (Heights*Width). We only use the last layer in each size-level and Relu function is applied to regulate the output of each layer. Let $S$ denote the size-level, then the output of the loss net is formulated as :

\begin{equation}
\phi_S = \mathcal{F}_{loss}(I)
\end{equation}

where $S$ is given a value range from 1 to 4, and each refers to the size-level 256,128,64,32. Each $\phi_S$ is a three-dimensional tensor describing the feature of the input $I$. And $\phi_S$ with higher level describes high-level semantic feature of style, while $\phi_S$ with lower level describes more detailed contents.

\subsection{Perceptual Loss Function}
The perceptual loss function was formulated to describe the tensor difference between $\hat{\phi_S}({I_{sCT}})$ and $\phi_S({I_{CT}})$. Since that Euclidean distance calculates an absolute matrix difference, it's commonly utilized as a "content" loss in \cite{li2016,tmi2018,tmi2020}, which is formulated as: 

\begin{equation}
\mathcal{L}_{content}(\hat{\phi_S},\phi_S) = \frac{1}{C*H*W}\sum_{c=1}^{C}\sum_{h=1}^{H}\sum_{w=1}^{W}||\hat{\phi_S} - \phi_S||_F^2
\end{equation}  

\par Using the absolute difference above as an optimization target can set strong spatial constraints between corresponding pixels and transfer more complete "content" information from the target domain to the input domain, but this absolute difference may incur an unstable training process. 

Gram matrix $\mathcal{G}(\phi)$ calculates the vector inner product of a tensor to describe a general value distribution style:  
\begin{equation}\label{vggselfloss}
\mathcal{G}(\phi) = \frac{1}{C*H*W}\sum_{h=1}^{H}\sum_{w=1}^{W}\phi_{h,w,c}\phi_{h,w,c^`}
\end{equation}

\par where $c^`$ refers to the transpose of matrix $\phi_{h,w,c^`}$ along "channel" axis. And the squared Frobenius norm between  $\mathcal{G}(\hat{\phi})$ and $\mathcal{G}(\phi)$ is utilized as a "style" loss\cite{li2016}:

\begin{equation}\label{vggloss}
\mathcal{L}_{style}(\hat{\phi},\phi) = ||\mathcal{G}(\hat{\phi}) - \mathcal{G}(\phi)||_F^2
\end{equation}  

\par The inner vector product measures cross-relation degrees between feature vectors, which enables a Gram matrix to describe a general value distribution style of a tensor. And the difference between Gram matrixes calculated from different tensors describes a "style" difference regardless of the specific spatial position of pixels. 

\par To ensure stable training and transfer comprehensive information, our proposed customed feature-to-feature perceptual(CFP)loss function is further formulated as:
\begin{equation}\label{perceptualloss}
\mathcal{L}_{CFP}(\hat{\phi},\phi) = \frac{a}{S_1}\sum_{1}^{S_1}\mathcal{L}_{content}(\hat{\phi_S},\phi_S) + \frac{b}{S_2}\sum_{1}^{S_2}\mathcal{L}_{style}(\hat{\phi},\phi)
\end{equation}

\par where $S_1$,$S_2$ refer to the size-level orders from 1 to 4(256 to 32), and they are customized as 2 and 4 in our case. $a$ and $b$ refer to weighing factors that are set as 0.5 and 0.5 through comparison experiments.

\section{Experiments and Results}

\subsection{Dataset}
In this study, we utilized four-dimensional thoracic CBCT and PCT image pairs from 100 lung cancer patients who underwent stereotactic radiotherapy on a Varian Medical Systems (VISION 3253) machine between 2017-2019 at Queen Mary’s Hospital in Hong Kong. Full-dose thin PCT slices (3.0mm) were acquired on a 16-row multi-detector helical CT scanner with a tube voltage of 120kV, variable tube current, variable exposure time, a gantry rotation time of 0.5s, a matrix size of 512 by 512 on the axial plane, and a pixel size of 1.074mm by 1.074mm. The full-dose thin CBCT slices (1.5mm-3.0mm) were acquired during SBRT treatment using the onboard imaging system on VISION 3253 with a tube voltage of 125kV, a tube current of 40 mA/frame, and variable exposure time. For each scan, a total of 360 projections were acquired in a full scan. All patients were placed in the Headfirst-Supine position during CBCT and PCT collection, following QMH's scanning protocol. PCT images were acquired only once about one to two weeks before the course of SBRT treatment. The average intensity projection was calculated for ten phases of each patient, resulting in 100 volume pairs of 4D-average CT and 4D-average CBCT for each patient. These 100 patients were randomly split 70/30 into AI-training and AI-testing groups, with the training dataset further split 56/14 for training and validation. 

\begin{figure*}[h]
  \center
  \includegraphics[width=7in]{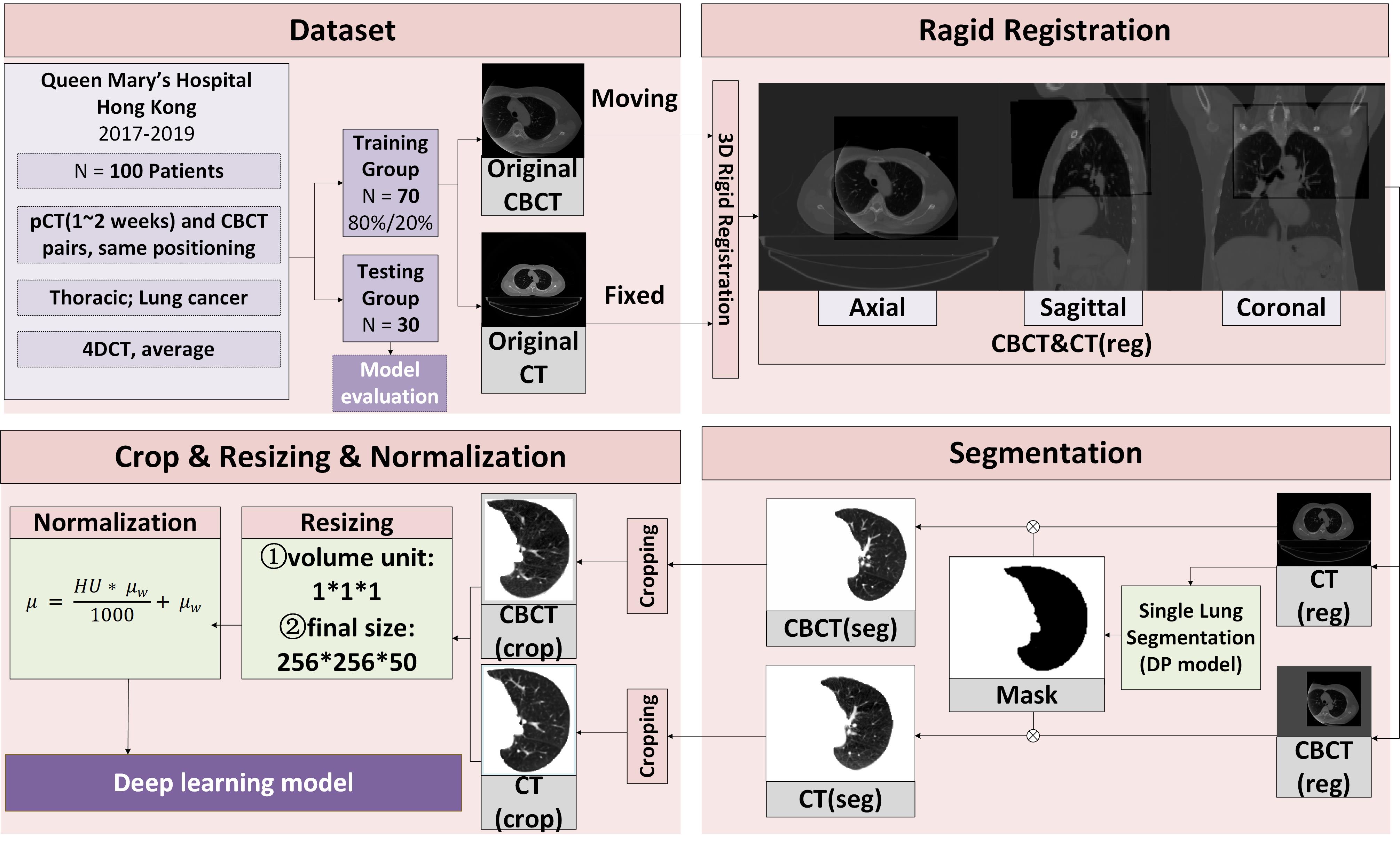}
  \caption{Preprocessing Workflow. Ragid registration, single lung segmentation, cropping,resizing and normalization are performed on the CT and CBCT images before they are fed into the deep learning network.}\label{preprocess}
\end{figure*}

\subsection{Preprocessing}

Fig.~\ref{preprocess} illustrates the preprocessing procedure used to obtain the training CBCT and CT images. To align the CBCT and CT images into the same coordinate system with a matrix size of 256x256x50, a 3D rigid registration was performed using the Elastix registration toolbox [51]. In order to focus the deep learning model on the pulmonary textual information and avoid the negative effects of non-anatomical structures during the training process, binary masks were created to separate the pulmonary region from the non-anatomical and surrounding thoracic regions. These masks were generated using the deep learning-based R231CovidWeb auto-segmentation model [52] on each CT and CBCT image from the same patient. The voxel values outside the mask region were replaced entirely with a Hounsfield Unit (HU) value of 0. The masked CT and CBCT images were then cropped and resized to remove any irrelevant background, and HU values outside the range of [-1000, 200] were cut off. Prior to training, all pixel values of the CT and CBCT images were transformed from HU values to linear attenuation coefficients (LAC) using the following formula:

\begin{equation}
    \mu = \frac{HU*\mu_w}{1000} + \mu_w
\end{equation}
where $\mu$ is the LAC and $\mu_w$ refers to the LAC of water,which is set to be 0.192. 

\subsection{Implement}
During the training process, we utilized the Adam optimization algorithm to optimize all component networks. The hyperparameters were set as follows: the learning rate was set to 1.0e-4, the batch size was set to 16, and the exponential decay rate was set to 0.9. We implemented the networks using PyTorch 1.10.1 and trained, tested, and validated them on a workstation equipped with an NVIDIA A6000 GPU.

\subsection{Quantitative Analysis}

We evaluated the imaging quality of our CBCT-to-CT translation model using five common metrics: peak-to-noise ratio (PSNR), structural similarity (SSIM), visual information fidelity (VIF), normalized cross-correlation (NCC), and information fidelity criterion (IFC). Mean values and deviations on the test dataset are reported in Table ~\ref{table2}. Of the five metrics, PSNR, SSIM, and NCC focus more on pixel-level similarity, while VIF and IFC take into account psychovisual features of the human visual system by using natural statistics models.

\begin{table*}[h]
\caption{The Quantitative test results of state-of-the-art methods with and without our proposed perceptual loss }
\centering
\setlength{\tabcolsep}{4.5mm}{
\begin{tabular}{cccccc} 
\hline\hline
\multirow{2}{*}{Method}   & \multirow{2}{*}{SSIM}            & \multirow{2}{*}{PSNR}             & \multirow{2}{*}{VIF}             & \multirow{2}{*}{IFC}             & \multirow{2}{*}{NCC}              \\ 
\\
\hline
Base                    & $0.8655_{ \pm 0.0435}$          & $28.6744_{ \pm 2.5962}$          & $0.1742_{ \pm 0.0645}$          & $1.1632_{ \pm 0.0111}$          & $0.3526_{ \pm 0.0967}$           \\ 

Unet-MSE                & $0.9327_{ \pm 0.0624}$          & $34.1525_{ \pm 2.1877}$          & $0.2004_{ \pm 0.0179}$          & $1.1125_{ \pm 0.0098}$          & $0.3324_{ \pm 0.0217}$           \\
Unet-CFP                & $0.9772_{ \pm 0.0227}$          & $\textbf{39.9621}_{ \pm 1.3511}$          & $\textbf{0.3021}_{ \pm 0.0108}$          & $\textbf{2.0015}_{ \pm 0.0073}$          & $0.3218_{ \pm 0.0189}$           \\ 

GAN                     & $0.9554_{ \pm 0.0525}$          & $37.1121_{ \pm 2.9887}$          & $0.2248_{ \pm 0.0493}$          & $1.4571_{ \pm 0.0207}$          & $0.3626_{ \pm 0.0913}$           \\
GAN-CFP                 & $0.9671_{ \pm 0.0326}$          & $38.8851_{ \pm 1.9561}$          & $0.2321_{ \pm 0.0366}$          & $1.7411_{ \pm 0.0117}$          & $0.4002_{ \pm 0.0897}$           \\ 

CycleGAN                & $0.9811_{ \pm 0.0497}$          & $37.2215_{ \pm 2.1564}$          & $0.2102_{ \pm 0.0343}$          & $1.4436_{ \pm 0.0169}$          & $0.4557_{ \pm 0.0981}$           \\
CycleGAN-CFP            & $\textbf{0.9869}_{ \pm 0.0298}$ & $39.1871_{ \pm 1.8003}$          & $0.2574_{ \pm 0.2333}$          & $1.5568_{ \pm 0.0128}$          & $\textbf{0.4971}_{ \pm 0.0910}$  \\
\hline\hline
\end{tabular}
}
\label{table1}
\end{table*} 

\subsection{Ablation Study}
In this section, we performed model ablation studies to evaluate our proposed multi-task perceptual loss customizing strategy. During the ablation studies, we gradually added subtasks to the multi-task perceptual loss customizing network and selected autoencoders from each trained network to build various perceptual loss functions. We used U-Net as the common CBCT-to-CT translation network and evaluated the generated sCT by U-Net with MSE and various perceptual loss functions quantitatively.The results are summarized in Table.~\ref{table1}, where the SSIM, PSNR, VIF, IFC and NCC were used to evaluate the similarity between CT targets and generated sCT images from CBCT inputs. The mean SSIM and PSNR values, along with deviations on the test dataset comprising 30 patients and a total of 3930 CBCT-CT slice pairs, were recorded in Table.~\ref{table2}.

\begin{table*}[h]
\caption{The Quantitative test results during Ablation study}
\centering
\setlength{\tabcolsep}{3.25mm}{
\begin{tabular}{cccccccc} 
\hline\hline
\multicolumn{3}{c}{\multirow{2}{*}{Loss Function Name}} & \multirow{2}{*}{SSIM}    & \multirow{2}{*}{PSNR}  & \multirow{2}{*}{VIF}   & \multirow{2}{*}{IFC}             & \multirow{2}{*}{NCC} \\ 
\\
\hline
\multicolumn{3}{c}{CBCT}   & $0.8655_{ \pm 0.0435}$  & $28.6744_{ \pm 2.5962}$ &$0.1742_{ \pm 0.0645}$ & $1.1632_{ \pm 0.0111}$ & $0.3526_{ \pm 0.0967}$\\
\multicolumn{3}{c}{MSE Loss}  & $0.9327_{ \pm0.0624}$     &  $34.1525_{ \pm 2.1877}$ & $0.2004_{ \pm 0.0179}$ & $1.1252_{ \pm 0.0098}$  & $0.3324_{ \pm0.0217}$       \\
\multicolumn{3}{c}{VGG Loss}  & $0.9206_{ \pm 0.0779}$     &  $35.1616_{ \pm 2.2115}$ & $0.2116_{ \pm 0.0321}$ & $1.1108_{ \pm 0.0079}$ & $0.3307_{ \pm 0.0169}$         \\ 
\hline
\multicolumn{3}{c}{CFP Loss}   &                    &                         \\
task 1  & task 2 & task 3      &                    &                          \\
\textbf{\checkmark} & & &  $0.9557_{ \pm 0.0197}$ & $36.7461_{ \pm 1.1778}$&$0.1939_{ \pm 0.0099}$ & $1.1539_{ \pm 0.0057}$ & $0.3501_{ \pm 0.0138}$  \\
           & \checkmark  & & $0.9151_{ \pm  0.0954}$&$32.1152_{ \pm 3.1152}$&$0.0917_{ \pm 0.0323}$ & $1.1413_{ \pm 0.0109}$ & $0.2899_{ \pm 0.0238}$            \\
           &   & \checkmark & $0.9254_{ \pm 0.0741}$ &  $28.2245_{ \pm 2.6612}$&$0.1067_{ \pm 0.0286}$ & $1.1517_{ \pm 0.0153}$ & $0.3105_{ \pm 0.0207}$          \\
\checkmark   & \checkmark  &  &   $0.9679_{ \pm 0.0256}$ & $37.1126_{ \pm 1.9845}$&$0.2501_{ \pm 0.0207}$ & $1.1610_{ \pm 0.0152}$ & $0.3216_{ \pm 0.0303}$       \\
 & \checkmark      & \checkmark   & $0.8517_{ \pm 0.0279}$&$ 27.2215_{ \pm 2.7745}$&$0.1058_{ \pm 0.0469}$ & $1.1208_{ \pm 0.0099}$ & $0.2915_{ \pm 0.0207}$     \\
\checkmark    &   & \checkmark    &  $0.9588_{ \pm 0.2928}$&  $38.7112_{ \pm 1.7511}$&$0.2479_{ \pm 0.0154}$ & $1.6172_{ \pm 0.0162}$ & $0.3311_{ \pm 0.0189}$           \\
\checkmark & \checkmark & \checkmark & $\textbf{0.9772}_{ \pm 0.0227}$& $\textbf{39.9621}_{ \pm 1.3551}$&$\textbf{0.3021}_{ \pm 0.0108}$ & $\textbf{2.0015}_{ \pm 0.0073}$ & $\textbf{0.3218}_{ \pm 0.0089}$ \\
\hline\hline
\end{tabular}
}
\label{table2}
\end{table*}

\subsection{Translation Results Comparison}
  In order to assess the performance of the CBCT-to-CT translation, we selected a representative case from the test set containing a complete left lung structure, a lung cancer lesion, and detailed airways and vessel anatomy. We generated sCT results using four typical generative networks, including Unet-MSE, Unet-CFP, GAN, GAN-CFP, CycleGAN, and CycleGAN-CFP, with and without our proposed CFP loss. We set the parameters of each generative model based on the settings of previous state-of-the-art CBCT-to-CT methods, including U-Net for \cite{uneten}, GAN for \cite{ganen} and CycleGAN for \cite{cycleganbrain}, \cite{cycleganprostate} and \cite{cycleganlung}. For GAN models, the "CFP" suffix refers to the replacement of the pixel-to-pixel MSE loss with feature-to-feature CFP perceptual loss.
  
\begin{figure*}
  \center
  \includegraphics[width=7in]{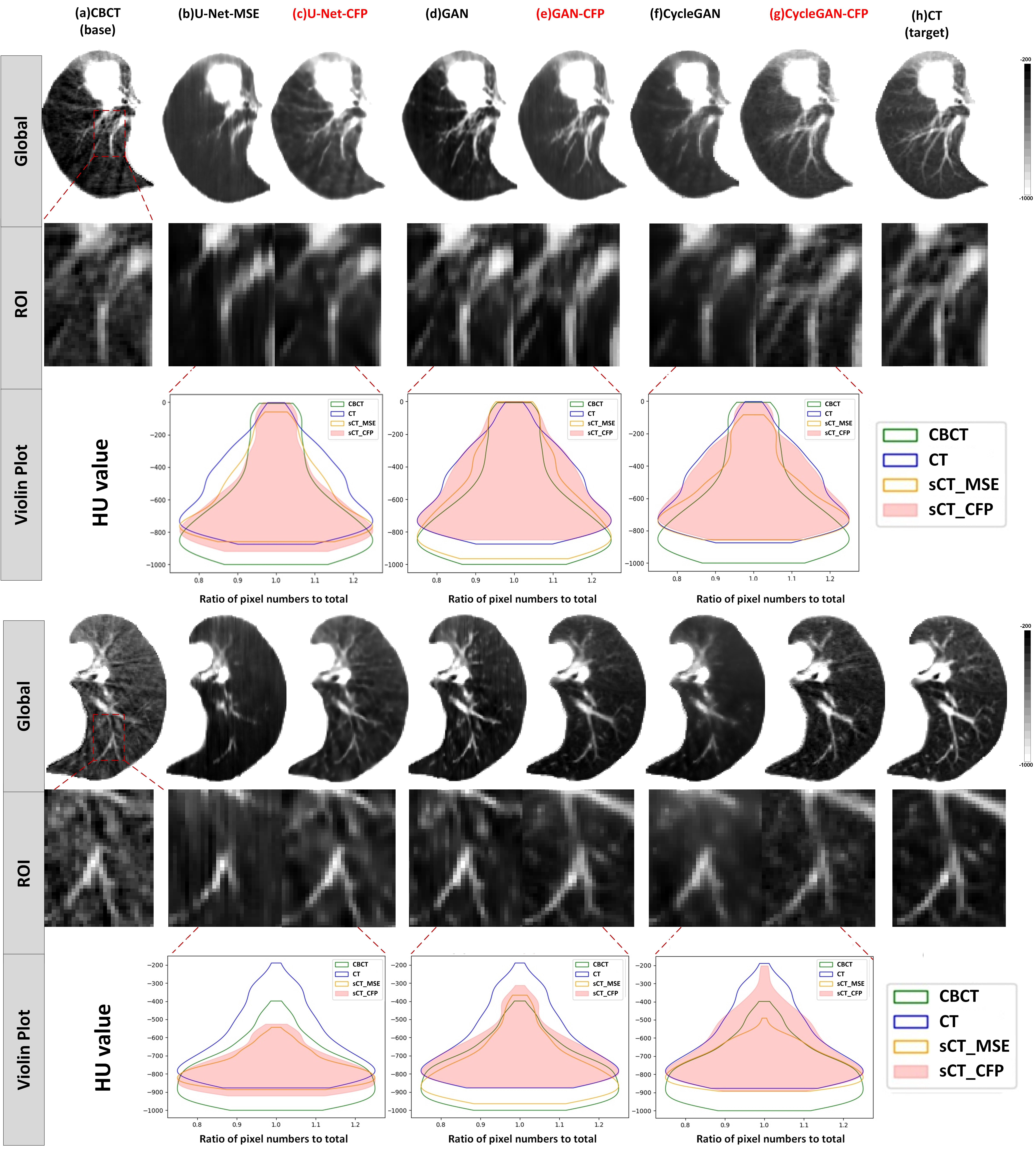}
  \caption{Visual performance comparison of different methods on a representative case of left lung and right lung. The CT display window is [-1000,-200]HU.}\label{roi}.
\end{figure*}

\begin{figure*}
  \center
  \includegraphics[width=7in]{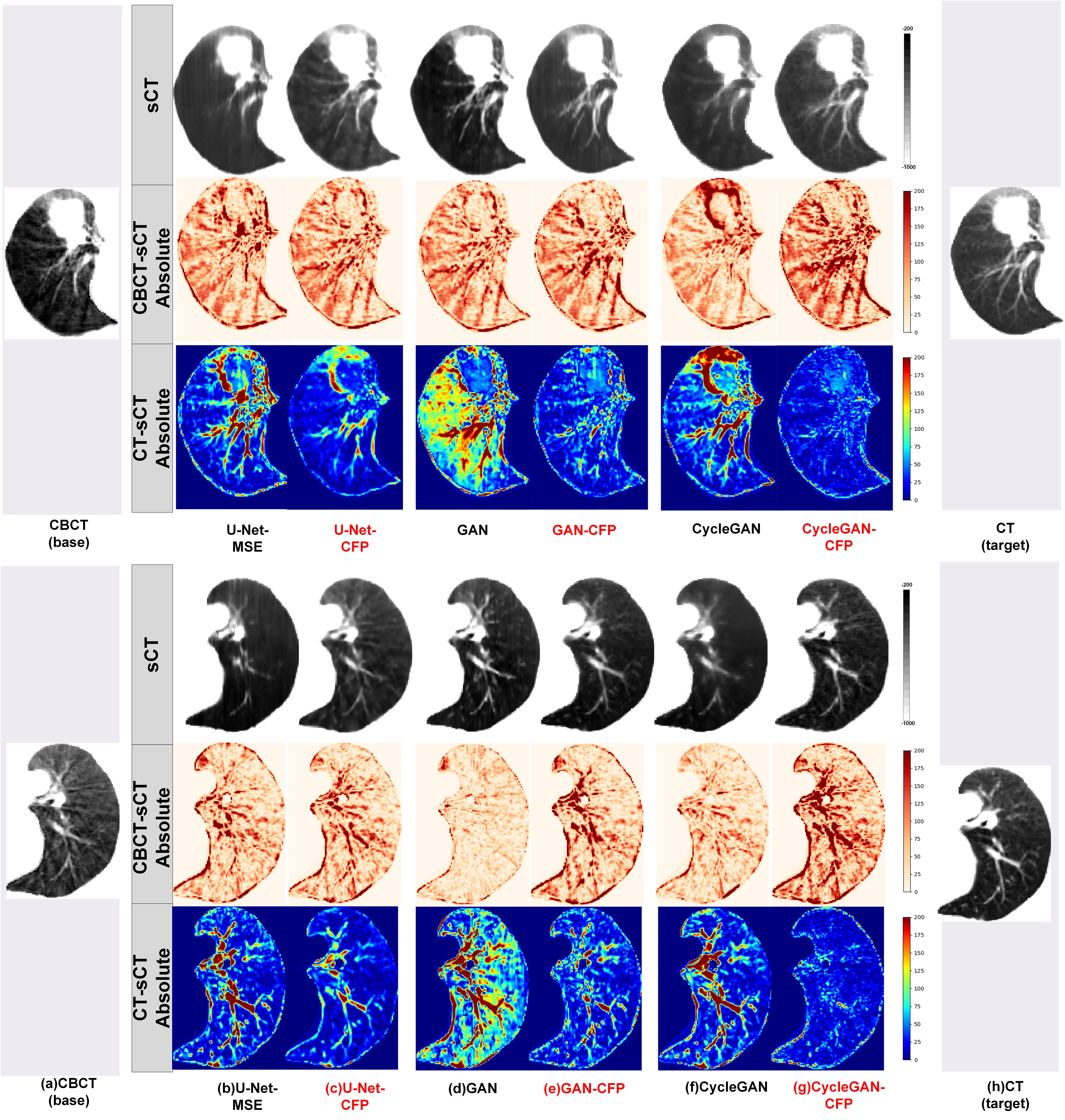}
  \caption{Absolute maps comparison of different methods on a representative case(CBCT-sCT/CT-sCT) of left lung and right. The CT display window is [-1000,-200]HU.}\label{ab}.
\end{figure*}

  Fig.~\ref{roi} presents the global visualization results, with the enlarged regions of interest (ROI) marked by dashed boxes, and violin curves on the representative case. Fig.~\ref{ab} shows the absolute maps between sCT and CBCT, and between sCT and CT.
  
\subsection{Functional Imaging Analysis}
\par
\begin{figure*}[h]
  \center
  \includegraphics[width=7in]{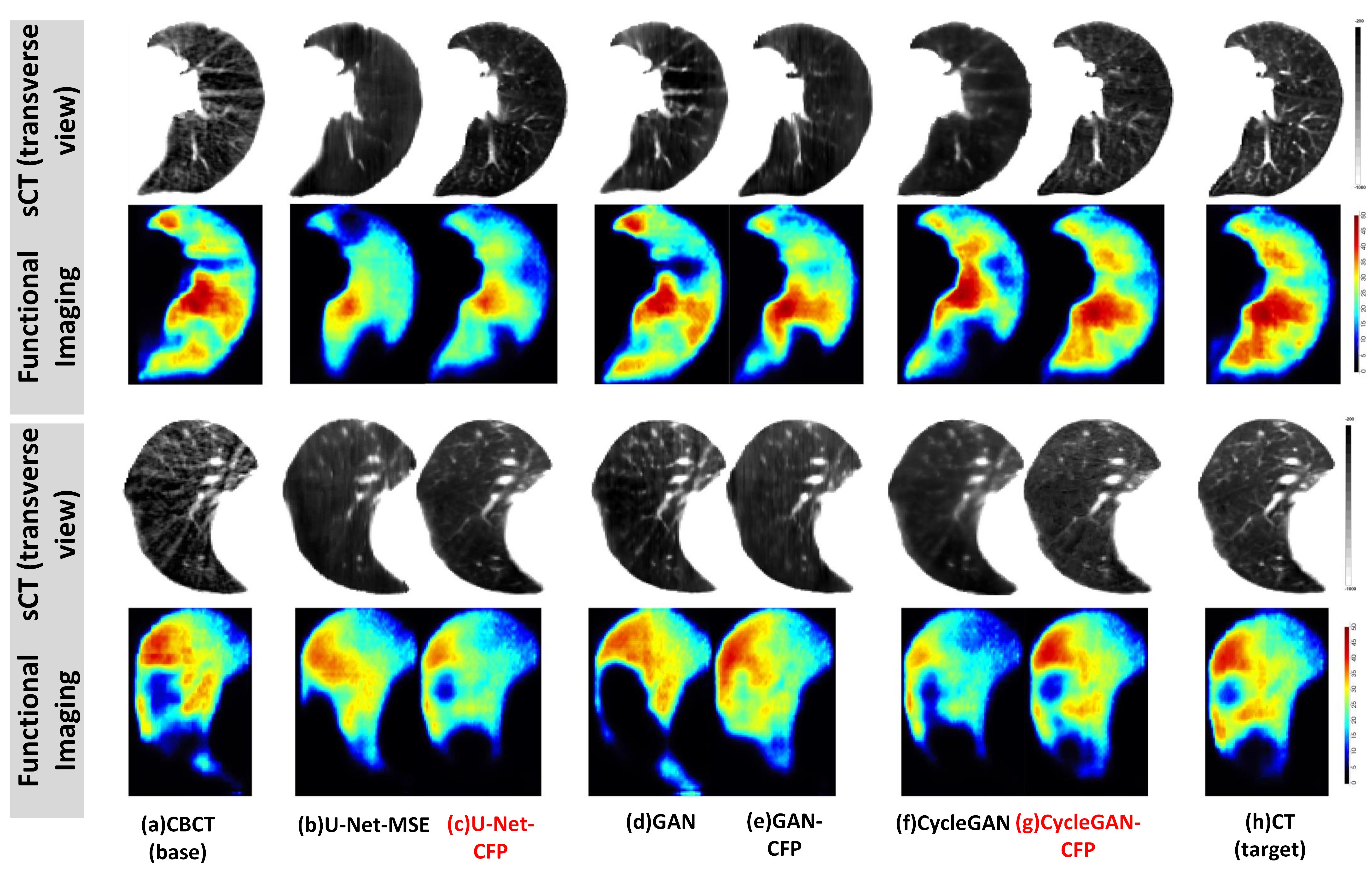}
  \caption{Functional imaging of four representative cases. The CT display window is [-1000,-200]HU.}\label{functional}
\end{figure*}

 To evaluate the clinical efficacy of CBCT imaging quality enhancement techniques, it is crucial to assess the HU value distribution. In order to evaluate the HU value correction performance of our method and its ability to preserve anatomical information, we applied a pulmonary perfusion generation model\cite{perfusion}  and analyzed the prediction accuracy of functional regions between CBCT, sCT, and CT pairs. We selected four representative cases, two with left lungs and two with right lungs, and the visual results are shown in Fig.~\ref{functional}.

\par To quantify the accuracy of functional region prediction, we calculated the Dice similarity coefficient (DSC) and Pearson's correlation coefficient (R) on the predicted functional imaging in the test dataset. The results are summarized in Table III. Among the five evaluation metrics, PSNR, SSIM, and NCC focus on pixel-level similarity, while VIF and IFC focus more on psychovisual features of the human visual system using natural statistics models.

To quantify the functional regions prediction accuracy, we also calculate Dice similarity correlation(DSC) and Pearsman's coefficient(R) on the predicted functional imaging on the test dataset. The results are summarized in Table.~\ref{table3}.

\begin{table}
\caption{The Quantative test results of functional imaging}
\centering
\setlength{\tabcolsep}{1mm}{
\begin{tabular}{cccc} 
\hline \hline
\multirow{2}{*}{Method} & \multirow{2}{*}{DSC}    & \multirow{2}{*}{SCC}    & \multirow{2}{*}{R}                  \\ 
 \\ 
\hline
Base                    & $0.7955_{ \pm 0.0611}$ & $0.9064_{ \pm 0.1007}$ & $0.9071_{ \pm 0.0753}$             \\ 

Unet-MSE                & $0.8708_{ \pm 0.0457}$ & $0.9093_{ \pm 0.0958}$ & $0.9174_{ \pm 0.0777}$             \\
Unet-CFP                & $0.8959_{ \pm 0.0172}$ & $0.9465_{ \pm 0.0413}$ & $0.9507_{ \pm 0.0698}$             \\ 

GAN                     & $0.7608_{ \pm 0.0391}$ & $0.7991_{ \pm 0.0823}$ & $0.8107_{ \pm 0.0686}$             \\
GAN-CFP                 & $0.8853_{ \pm 0.0128}$ & $0.9284_{ \pm 0.0625}$ & $0.9343_{ \pm 0.0574}$             \\ 

CycleGAN                & $0.8299_{ \pm 0.0351}$ & $0.8705_{ \pm 0.0653}$ & $0.8801_{ \pm 0.0596}$             \\
CycleGAN-CFP            & $\textbf{0.9147}_{ \pm 0.0086}$ & $\textbf{0.9615}_{ \pm 0.0322}$ & $\textbf{0.9661}_{ \pm 0.0399}$             \\
\hline \hline
\end{tabular}
}
\label{table3}
\end{table}

\section{Discussion}
\par In this study, we developed a novel deep learning framework for enhancing lung CBCT imaging quality to CT level, which consists of two main components: a multi-task feature-selection network and a CBCT-to-CT translation network. The multi-task feature-selection network is designed to pretrain a customized autoencoder that serves as a perceptual loss building-up network with fixed parameters in the CBCT-to-CT translation network. The autoencoder is customized with mixed traits by jointly training three unique tasks, and the loss functions from these tasks are united through a gradnorm regularization technique to further refine the autoencoder. The CBCT-to-CT translation network generates high-quality synthesized CT (sCT) imaging from CBCT by using a perceptual loss between the sCT and paired CT images. This perceptual loss is defined by the difference between level-crossing feature maps, which are extracted by different layers from the perceptual loss network. The multi-task feature-selection network is responsible for building up this network. 

\par Thoracic CBCT can be collected conveniently during lung radiation therapy with real-time anatomy and is potentially qualified for cancer-based positioning, adaptive treatment planning , radionics analysis, and lung functional imaging synthesis. High-quality thoracic CBCT is also a precondition of real-time treatment planning. However, other than the severe artifacts, noise, low soft tissue contrast and HU value inconsistency, the poor reservation of tiny and detailed pulmonary anatomy can be the most challenging factor hampering the clinical application of CBCT. Our proposed feature-oriented framework provides a solution for pulmonary fine anatomy reservation.

\par The proposed feature-oriented framework presents a significant advance over existing CNN and GAN based models for CBCT-to-CT translation, which have typically focused on the receptive field, network structure modification, or pairs match. The pixel-to-pixel loss functions employed in these approaches are inadequate for comprehensively describing the differences between modalities, spatial locations, and anatomical structures in a mismatch scenario like CBCT-to-CT translation. In contrast, our framework leverages a perceptual loss that comprehensively captures the high-dimensional feature map differences between paired CBCT and CT images, enabling the transfer of noise-free and clear anatomical styles from CT to CBCT at a perceptual level .

nor limited to a simple filtering function

\par The sCT images generated by our proposed feature-oriented framework show great imaging quality improvement compared to most of the current pixel-to-pixel generative models. In terms of statistical similarity between sCT and CT, our proposed method achieved the highest SSIM, PSNR, VIF, IFC and NCC compared to pixel-to-pixel Unet,GAN or CycleGAN models. The dominant quantitative results demonstrate great improvement of anatomical information preservation ability owing to our proposed feature-level optimization target. In terms of qualitative analysis, compared to other methods, sCT images generated by our framework show superior artifacts suppression, noise reduction and clear anatomical details both globally and locally. The great visual performance proves that our proposed multi-task feature-selection strategy has great CBCT-to-CT feature extraction ability and is an effective solution to the CBCT-to-CT translation scenario. Ablation experiment further proves that the self-recovery, registration and classification tasks are complementary and together build up a performant feature-selection network. In the lung function imaging synthesis experiment, compared to other methods, our framework shows more precise high-function regions prediction.
These results in-depth demonstrate the HU value distribution correction ability of our proposed framework. 

\par To the best of our knowledge, our proposed framework is the first post-processing based CNN for CBCT refinement that operates in a feature-to-feature level. It is especially effective for enhancing pulmonary CBCT images with a high level of detail. The loss functions used in our framework set strong constraints on the mapping between input and target spaces, and all network modification operations share the common goal of fitting an appropriate function to achieve the desired mapping pattern within the assumption space. However, further research is needed to explore the limitations of network architecture modifications such as attention, training strategy, or generative models in this context:

customize the output layers

\begin{enumerate}
\item Expand our CBCT imaging quality enhancement framework to fit more body regions: Essentially, the severe streaking artifacts in CBCT images are caused by information loss that varies among different tissues or organs. Therefore, the characteristics of CBCT streaking artifacts may vary across different body regions. To address this, we plan to expand our pulmonary CBCT imaging quality enhancement technique to the whole thoracic region. We aim to train separate customized perceptual loss functions and translation networks for enhancing the segmented pulmonary region and the rest of the thoracic region. Additionally, we plan to test our framework with datasets covering other body regions, such as the head and neck region or pelvic region.

\item  Find a better evaluation method: Metrics such as SSIM or PSNR are designed to evaluate pixel-to-pixel similarity. However, in the case of CBCT-to-CT translation, where the CBCT and CT pairs are mismatched and contain misaligned anatomical information, a pixel-to-pixel similarity metric may be inadequate to comprehensively describe the anatomy preservation performance of the synthesized CT (sCT) images from CBCT. Therefore, we plan to explore high-level correlations between CBCT and sCT and develop a more suitable evaluation method to quantify the anatomy preservation performance of sCT images from CBCT.

\item Focus more on HU value correction: Our multi-task feature-selection network is designed to transfer CT-like traits from a CT self-recovery task to the autoencoder, which serves as a feature extraction loss network. The perceptual loss function is directly calculated from the feature maps extracted by this network. While our framework has demonstrated an overall ability to correct Hounsfield Unit (HU) value distributions, we need to provide more detailed HU value distribution differences between CBCT and CT in our scheme. Firstly, a further manual evaluation of HU values should be included, considering a clinical HU range for different tissues and organs. Secondly, we plan to focus more on precise HU value correction by quantifying the HU distribution differences between CT and CBCT modalities.

\end{enumerate}

\section{Conclusion}
In this study, we developed a novel deep learning framework for improving the quality of pulmonary CBCT imaging to CT level. The multi-task feature-selection network is designed to pretrain a customized autoencoder that serves as a perceptual loss building-up network with fixed parameters in the CBCT-to-CT translation network. Our proposed framework outperforms other state-of-the-art methods by generating sCT images with superior imaging quality. Quantitative and qualitative experiments show that the sCT images achieve superior fine anatomical details preservation and artifact suppression. In addition, functional imaging analysis further demonstrates the HU value correction performance and clinical efficacy of our framework.

\section*{Acknowledgements}
This research was partly supported by General Research Fund (15103520), the University Research Committee, Health and Medical Research Fund (07183266, 09200576), PolyU (UGC) Start-up Fund for RAPs under the Strategic Hiring Scheme (P0038378), PolyU (UGC) RI-IWEAR Seed Project (P0044802), the Health Bureau, The Government of the Hong Kong Special Administrative Region.

\bibliographystyle{IEEEtran} 
\bibliography{reference}

\end{document}